\documentclass[letterpaper, 10 pt, journal, twoside]{./IEEEtran}
\usepackage{graphicx}
\usepackage{color}
\usepackage{cite}
\usepackage[hidelinks,bookmarks=false]{hyperref}
\usepackage{multirow}
\usepackage{enumerate}
\usepackage{bm}
\usepackage{subfigure}
\usepackage{booktabs}
\usepackage{blindtext}
\usepackage[T1]{fontenc}
\usepackage{mathptmx} 
\DeclareMathAlphabet{\mathcal}{OMS}{cmsy}{m}{n}
\usepackage{times} 
\usepackage{amsmath} 
\usepackage{amssymb}  
\usepackage[ruled,linesnumbered]{algorithm2e}
\usepackage{url}
\usepackage{bbm}
\usepackage{booktabs}
\usepackage{upgreek}

\ifCLASSINFOpdf
\else
\fi
\begin{document}
\makeatletter
\newcommand{\rmnum}[1]{\romannumeral #1}
\newcommand{\Rmnum}[1]{\expandafter\@slowromancap\romannumeral #1@}
\makeatother
%
\title{High-speed Autonomous Drifting with Deep Reinforcement Learning}
%
%
%

\author{Peide~Cai$^{*1}$, Xiaodong~Mei$^{*1}$, Lei~Tai$^{2}$, Yuxiang~Sun$^{1}$, and Ming~Liu$^{1}$%
\thanks{*The first two authors contributed equally to this work.} 
\thanks{$^{1}$P. Cai, X. Mei, Y. Sun and M. Liu are with The Hong Kong University of Science and Technology (email: {pcaiaa, xmeiab, eeyxsun, eelium}@ust.hk).
        }%
\thanks{$^{2}$L. Tai is with Alibaba Group (email: tailei.tl@alibaba-inc.com).
        }%
}
\maketitle

\begin{abstract}
Drifting is a complicated task for autonomous vehicle control. Most traditional methods in this area are based on motion equations derived by the understanding of vehicle dynamics, which is difficult to be modeled precisely. We propose a robust drift controller without explicit motion equations, which is based on the latest model-free deep reinforcement learning algorithm soft actor-critic. The drift control problem is formulated as a trajectory following task, where the error-based state and reward are designed. After being trained on tracks with different levels of difficulty, our controller is capable of making the vehicle drift through various sharp corners quickly and stably in the unseen map. The proposed controller is further shown to have excellent generalization ability, which can directly handle unseen vehicle types with different physical properties, such as mass, tire friction, etc.
\end{abstract}


%
\IEEEpeerreviewmaketitle

\section{Introduction}
\begin{figure}[t]
    \centering
    \setlength{\abovecaptionskip}{-1pt}
    \includegraphics[width = \columnwidth]{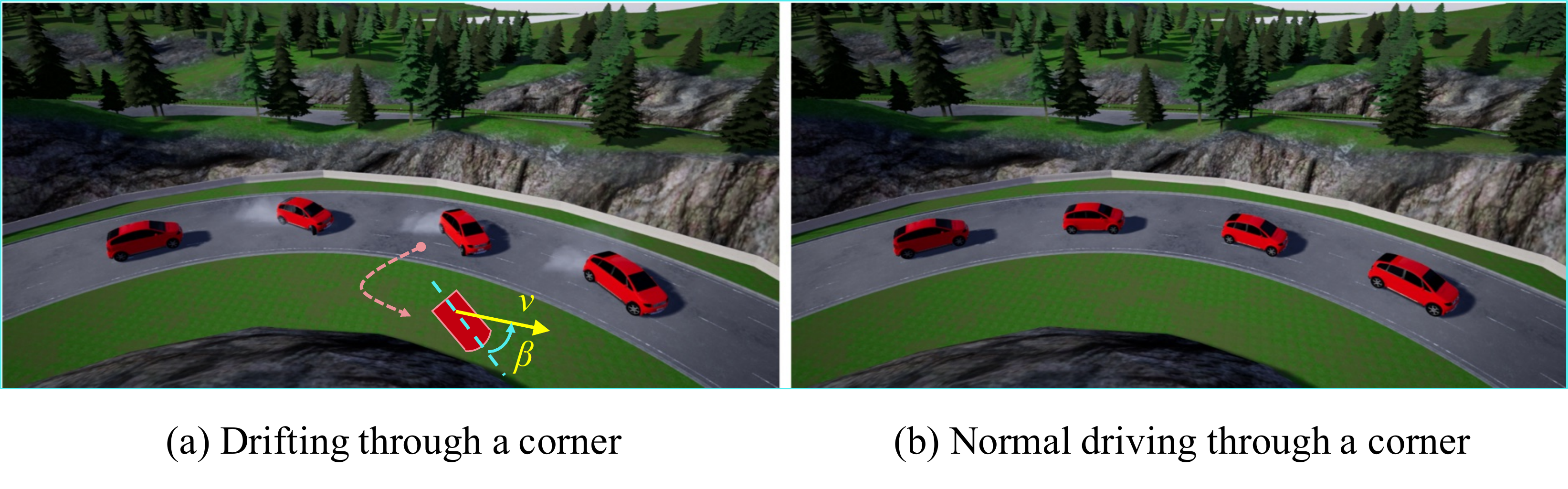}
    \caption{Comparison between drifting and normal driving through a corner. A drift car usually has a large slip angle $\beta$ with saturated rear tires caused by oversteering, which is often evidenced by large amounts of tire smoke.}
    \label{drift_and_normal}
    \vspace{-0.3cm} 
\end{figure}

%
%
%
%
\IEEEPARstart{I}{n} motorsport of rallying, high-speed sideslip cornering, known as drifting, represents an attractive vehicle control maneuver undertaken by professional racing drivers. The slip angle $\beta$ is measured by the angle between the direction of the heading (longitudinal axis of the vehicle) and the direction of the velocity vector at the centre of gravity, as shown in Fig. \ref{drift_and_normal}(a). In order to make a quick turn through sharp corners, skilled drivers execute drifts by deliberately inducing deep saturation of the rear tires by oversteering\cite{zubov2018autonomous} or using the throttle\cite{voser2010analysis}, thereby destabilising the vehicle. They then stabilise the vehicle as it begins to spin by controlling it under a high sideslip configuration (up to 40 degrees\cite{acosta2018teaching}). Vehicle instability and corresponding control difficulty both increase as the sideslip angle increases. Therefore, drifting is a challenging control technique to operate the vehicle efficiently and safely beyond its handling limits. Compared with the normal cornering in which slipping is usually avoided by lowering the speed and making gentle turns (Fig. \ref{drift_and_normal}(b)), high-speed drifting techniques can help reduce the lap time during racing \cite{zhang2017autonomous,zhang2018drift,bhattacharjee2018autonomous,frere2016sports}.

The fact that racing drivers deliberately drift through sharp corners indicates that there is a lot of knowledge about agile control to be learned. During drifting, a series of high-frequency decisions like steering and throttle should be executed precisely and safely. Therefore, by studying drift behaviors, we can design controllers which fully exploit vehicle dynamics to reduce lap time with high-speed sideslip cornering for racing games. The results could further contribute to the understanding of aggressive driving techniques and extend the operating envelope for autonomous vehicles.

Most of the previous works on drift control are based on the understanding of vehicle dynamics \cite{ drift8, goh2016simultaneous, velenis2011steady,velenis2005minimum, hindiyeh2014controller }, including tire forces and moments generated by the wheels. Then models with varying fidelities and explicit motion equations are utilized to develop the required controllers by classical, adaptive or optimal control methods. However, in these methods, tire parameters such as longitudinal stiffness at different operating points have to be identified in advance, which is extremely complicated and costly \cite{acosta2018teaching}. It is also not easy to accurately derive the entire vehicle dynamics, because some parts of the system are hard to model, and exceeding the handling limits of these models could lead to strong input coupling and sideslip instability \cite{drift8}. 

\begin{figure*}[t]
    \centering
    \includegraphics[width = 2\columnwidth]{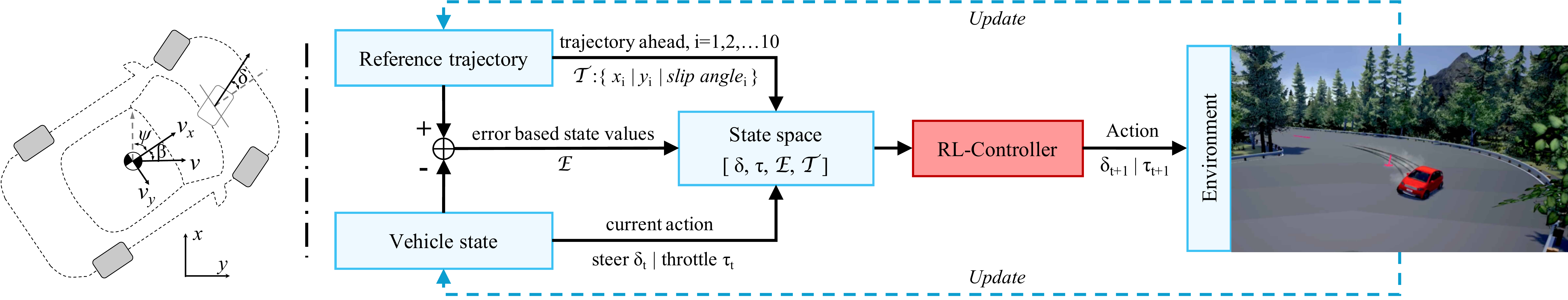}
    \caption{State variables of the vehicle (left) and the control loop of our deep RL-based method (right). The steering angle is $\delta$. The heading angle $\psi$ is defined as the angle between the direction of the heading and the direction of the world-frame x. The forward and side velocities of the vehicle are $v_x$ and $v_y$ respectively, with $v$ being the total velocity. The angle between the direction of the heading and the direction of $v$ is called the slip angle $\beta$. For the control loop, the deep RL-Controller receives observations from the neighboring reference trajectory and the vehicle state. Then it produces an action composed of the steering angle and throttle to operate the simulated car. Finally, the environment feeds back the updated vehicle state and the reference trajectory, to be utilized in the next control step.}
    \label{control_loop}
    \vspace{-0.3cm}
\end{figure*}

The aforementioned limitations motivate the exploration of strategies to agilely control the drifting vehicles without tire models or explicit motion equations. It is a perfect use case for the learning-based methods, especially model-free deep reinforcement learning (RL). Instead of relying on the human understanding of the world to design controllers, model-free deep RL methods learn the optimal policy by \textit{interacting} with the environment. Prior learning-based works on autonomous drifting \cite{bhattacharjee2018autonomous, cutler2016autonomous} mostly consider \textit{sustained} drift by stabilizing the vehicle states about a single drift equilibrium (e.g., steady state circular drift), which is straightforward but not practical. Thus, a novel learning-based method to realize high-speed \textit{transient} drift by tracking a set of non-steady drift states (e.g., drift cornering) is discussed in this paper. The main contributions of this paper are as follows.

\begin{itemize}
    \item We design a closed-loop controller based on model-free deep RL to control front-wheel drive (FWD) vehicles to drive at high speed (80-128 $km/h$), and to drift through sharp corners quickly and stably following a reference trajectory, as shown in Fig. \ref{control_loop}.
    \item We evaluate the proposed controller on various environmental configurations (corner shapes, vehicle types/mass and tire friction) and show its notable generalization ability.
    \item We open source our code for benchmark tests and present a dataset for future studies on autonomous drifting. The dataset contains seven racing maps with reference drift trajectories. \footnote{https://sites.google.com/view/autonomous-drifting-with-drl/}
\end{itemize}

\section{Related Work}
\subsection{Reinforcement learning algorithms}
Reinforcement learning is an area of machine learning concerning how agents should take actions to maximize the sum of expected future rewards. The action ($\mathbf{a}_{t}$) is taken according to a policy $\pi: \mathbf{s}_t \rightarrow \mathbf{a}_t$, where $\textbf{s}_t$ is the current state. The policy is then evaluated and updated through repeated interactions with the environment by observing the next state ($\mathbf{s}_{t+1}$) and the received reward ($r_t$).

RL algorithms are divided into model-based and model-free types. Different from model-based RL algorithms such as probabilistic inference for learning control (P\textsc{ilco}) \cite{deisenroth2011pilco}, model-free RL eliminates the complex and costly modeling process entirely. Combined with deep neural networks as nonlinear function approximators, model-free RL has been applied to various challenging areas. The algorithms can be divided into value-based and policy gradient algorithms. Value-based methods, such as DQN\cite{mnih2015human}, learn the state (or action) value function and select the best action from a discrete space, while policy gradient methods directly learn the optimal policy, which extend to a continuous action space. The actor-critic framework is widely used in policy gradient methods. Based on this framework, Lillicrap et al.\cite{lillicrap2015continuous} propose deep deterministic policy gradients (DDPG) with an off-policy learning strategy, where the previous experience can be used with a memory replay buffer for better sample efficiency. However, this method is difficult to converge due to the limited exploration ability caused by its deterministic character. To improve the convergence ability and avoid the high sample complexity, one of the leading state-of-the-art methods called soft actor-critic (SAC)\cite{haarnoja2018soft} is proposed. It learns a stochastic actor with an off-policy strategy, which ensures sufficient exploration and efficiency for complex tasks. 

\subsection{Drifting control approaches}

\subsubsection{Traditional methods}
Different levels of model fidelity depicting the vehicle dynamics have been used in prior works for the drift controller design. A two-state single-track model is used by Voser et al. \cite{voser2010analysis} to understand and control high sideslip drift maneuvers of road vehicles. Zubov et al. \cite{zubov2018autonomous} apply a more-refined three-state single-track model with tire parameters to realize a controller stabilizing the all-wheel drive (AWD) car around an equilibrium state in the Speed Dreams Simulator. 

Although these methods have been proposed to realize steady-state drift, transient drift is still an open problem for model-based methods, mainly due to the complex dynamics while drifting. Velenis et al. \cite{velenis2005minimum} introduce a bicycle model with suspension dynamics and apply different optimization cost functions to investigate drift cornering behaviors, which is validated in the simulation. For more complex trajectories, Goh et al. \cite{drift8} use the rotation rate for tracking the path and yaw acceleration for stabilizing the sideslip, and realize automated drifting along an 8-shaped trajectory.

These traditional drift control methods rely on the knowledge of tire or road forces, which cannot be known precisely due to the real-world environmental complexity. In addition, inaccuracies in these parameters will lead to poor control performance.

\begin{figure*}[t]
    \centering
    \setlength{\abovecaptionskip}{0pt}
    \includegraphics[width = 2\columnwidth]{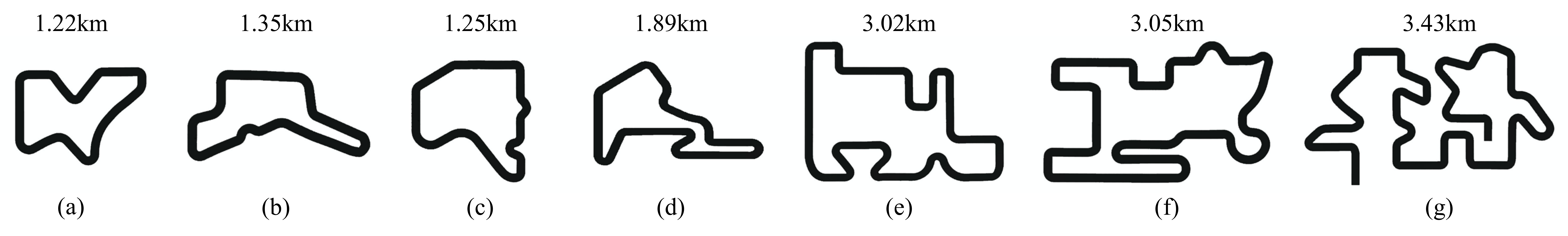}
    \caption{Seven maps are designed for the drifting task. The difficulty of driving increases from (a) to (g). (a-f) are for training and (g) is for evaluation.}
    \label{maps}
    \vspace{-0.3cm}
\end{figure*}

\subsubsection{Learning-based methods}
Cutler et al. \cite{cutler2016autonomous} introduce a framework that combines simple and complex simulators with a real-world remote-controlled car to realize a steady-state drift with constant sideways velocity, in which a model-based RL algorithm, P\textsc{ilco}, is adopted. Bhattacharjee et al. \cite{bhattacharjee2018autonomous} also utilize P\textsc{ilco} to realize sustained drift for a simple car in the Gazebo simulator. Acosta et al. \cite{acosta2018teaching} propose a hybrid structure formed by the model predictive controller (MPC) and neural networks (NNs) to achieve drifting along a wide range of road radii and slip angles in the simulation. The NNs are used to provide reference parameters (e.g., tire parameters) to the MPC, which are trained via supervised learning.

Our work differs from the aforementioned works in several major ways. First, we adopt SAC, the state-of-the-art model-free deep RL algorithm, to train a closed-loop drift controller. To the best of the authors knowledge, this is the first work to achieve transient drift with deep RL. Second, our drift controller generalizes well on various road structures, tire friction and vehicle types, which are key factors for controller design but have been neglected by prior works in this field.

\section{Methodology}
\subsection{Formulation}
We formulate the drift control problem as a trajectory following task. The goal is to control the vehicle to follow a trajectory at high speed (>80 $km/h$) and drift through manifold corners with large sideslip angles (>20$^{\circ}$), like a professional racing driver. We design our controller with SAC and use CARLA \cite{dosovitskiy2017carla} for training and validation. CARLA is an open-source simulator providing a high-fidelity dynamic world and different vehicles of realistic physics.

\subsubsection{Map generation}
Seven maps (Fig. \ref{maps}) with various levels of difficulty are designed for the drifting task, for which we refer to the tracks of a racing game named PopKart \cite{popkart}. These are generated by RoadRunner \cite{roadrunner}, a road and environment creation software for automotive simulation.

\subsubsection{Trajectory generation}

For a specific environment, we aim to provide our drift controller with a candidate trajectory to follow. However, the prior works from which to generate reference drift trajectories\cite{goh2016simultaneous, drift8} are based on simplified vehicle models, which are rough approximations of the real physics. To better train and evaluate our controller, more suitable trajectories are needed. To this end, we invite an experienced driver to operate the car with steering wheel and pedals (Logitech G920) on different maps and record the corresponding trajectories. The principle is to drive as fast as possible and use drift techniques for cornering sharp bends. The collected data contains the vehicle world location, heading angles, body-frame velocities and slip angles, to provide reference states for training and evaluation.

\subsection{RL-based drift controller}

\begin{figure}[tpb]
    \centering
    \setlength{\abovecaptionskip}{-0.3pt}
    \includegraphics[width = 0.75\columnwidth]{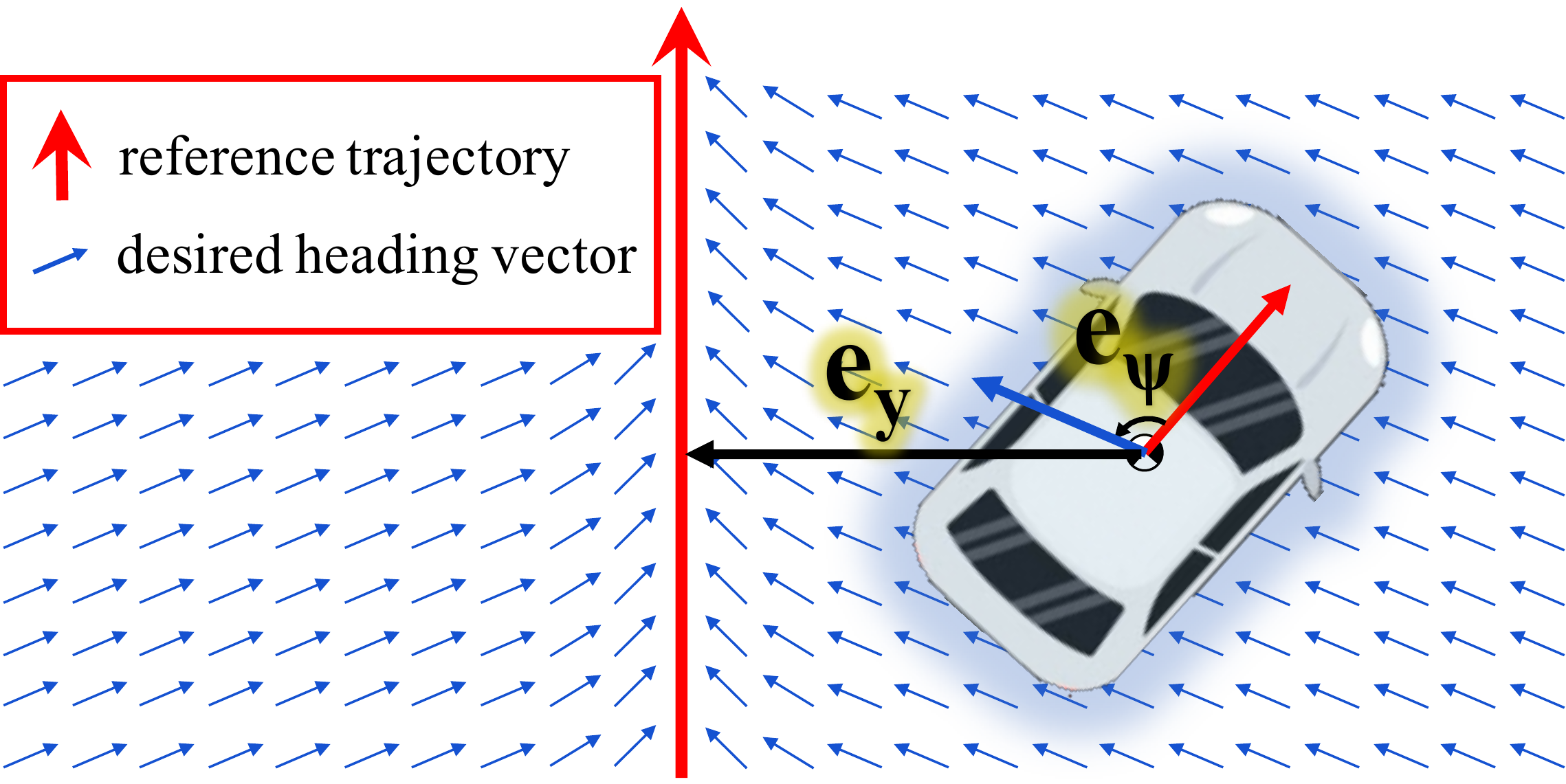}
    \caption{Vector field guidance (VFG) for drift control. $e_{y}$ is the cross track error, defined as the perpendicular distance of the vehicle from the reference track. $e_{\psi}$ is the heading angle error, which is the difference between the heading angle of the vehicle and the desired heading angle provided by VFG.}
    \label{vfg}
    \vspace{-0.3cm}
\end{figure}

\subsubsection{State variables}
The state variables of the vehicle include steering angle $\delta$, throttle $\tau$, forward and side velocities ($v_x$, $v_y$), total velocity $v$, side slip angle $\beta$ and heading angle $\psi$, as depicted in Fig. \ref{control_loop}. For an arbitrary location of the vehicle, we adopt the vector field guidance (VFG) \cite{nelson2007vector} to determine the desired heading angle $\psi^d$. Fig. \ref{vfg} demonstrates a VFG for a linear path and related error variables, which are cross track error $e_y$ and heading angle error $e_{\psi}$. The objective of the constructed vector field is that when $e_y$ is small, $\psi^d$ is close to the direction of the reference trajectory $\psi^{ref}$. As $e_y$ increases, their difference increases as well:
\begin{equation}
    \psi^{d} = d \psi^{\infty} \frac{2}{\pi} \text{tan}^{-1}\left( k e_y \right) + \psi^{ref},
\end{equation}
where $d = 1$ if the vehicle is on the west of the reference path, or else $d=-1$. $k$ is a positive constant that influences the rate of the transition from $\left( \psi^{ref} \pm \psi^{\infty} \right)$ to $\psi^{ref}$. Large values of $k$ result in short and abrupt transitions, while small values cause long and smooth transitions. In this work, we choose $k=0.1$. $\psi^{\infty}$ is the maximum deviation between $\psi^{d}$ and $\psi^{ref}$, which is set to 90$^\circ$.

\subsubsection{State space}
Based on the state variables introduced above, the state space $\mathbf{s} \in \mathcal{S}$ is defined as (\ref{state_space}),
\begin{equation}
\mathcal{S}=\left\{\delta,\tau,e_y,\dot e_y,e_{\psi},\dot e_{\psi},e_{\beta},\dot e_{\beta},e_{vx},
\dot e_{vx}, e_{vy}, \dot e_{vy}, \mathcal{T}\right\},
\label{state_space}
\end{equation}
where $\mathcal{T}$ contains ten $(x,y)$ positions and slip angles in the reference trajectory ahead. Therefore, the dimension of $\mathcal{S}$ is 42. $e_{\beta}$ is the slip angle difference between the vehicle and the reference trajectory. $e_{vx}$ and $e_{vy}$ is the error of the forward and side velocity, respectively. Moreover, time derivatives of the error variables, such as $\dot e_{y}$, are included to provide temporal information to the controller \cite{woo2019deep}. We also define the terminal state with an \textit{endFlag}. When the vehicle is in collision with barriers, arrives at the destination or is over fifteen meters away from the track, \textit{endFlag} becomes \textit{true} and the current state changes to terminal state $\textbf{s}_{T}$.

\subsubsection{Action space}
The continuous action space $\mathbf{a} \in \mathcal{A}$ is defined as (\ref{action_space}),
\begin{equation}
\mathcal{A}=\left\{\delta,\tau \right\}.
\label{action_space}
\end{equation}

\begin{algorithm}[t]
    \caption{SAC controller training algorithm \label{alg:sac}}
    \KwData{Buffer $\mathcal{D}$, total number of transitions N, update threshold $\eta$, number of updates $\lambda$}
    \KwResult{Optimal control policy $\pi^{\ast}_{\upphi}$}
    Initialize parameters of all networks.\\
    $\mathcal{D}\leftarrow \emptyset,\ N\leftarrow 0,\ \textbf{s}_t\leftarrow \textbf{s}_0$\;
    \For{each episode}
    {
    \While{$\textbf{s}_t \neq \textbf{s}_{T}$}{
    $\textbf{a}_t \thicksim \pi_{\upphi}(\textbf{a}_t\lvert\textbf{s}_t),\ \textbf{s}_{t+1} \thicksim \mathnormal{p}(\textbf{s}_{t+1}\lvert\textbf{s}_t,\textbf{a}_t)$\;
    $\mathcal{D}\leftarrow\mathcal{D} \bigcup \{(\textbf{s}_t,\textbf{a}_t,r(\textbf{s}_t,\textbf{a}_t),\textbf{s}_{t+1})\}$\;
    $N\leftarrow N+1,\ \textbf{s}_t \leftarrow \textbf{s}_{t+1}$\;}
    \If{$N \geqslant \eta$}
    {Update all networks $\lambda$ times\;}
    }
\end{algorithm}

In CARLA, the steering angle $\delta$ and throttle $\tau$ is normalized to $[-1,1]$ and $[0,1]$, respectively. Since the vehicle is expected to drive at high speed, we further limit the range of the throttle to $[0.6,1]$ to prevent slow driving and improve training efficiency. Additionally, according to the control test in CARLA, high-speed vehicles are prone to rollover if large steering angles are applied. Therefore, the steering is limited to a smaller range of $[-0.8,0.8]$ to prevent rollover.

\begin{figure}[t]
    \centering
    \vspace{-0.3cm}
    \setlength{\abovecaptionskip}{-3pt}
    \includegraphics[width = 0.85\columnwidth]{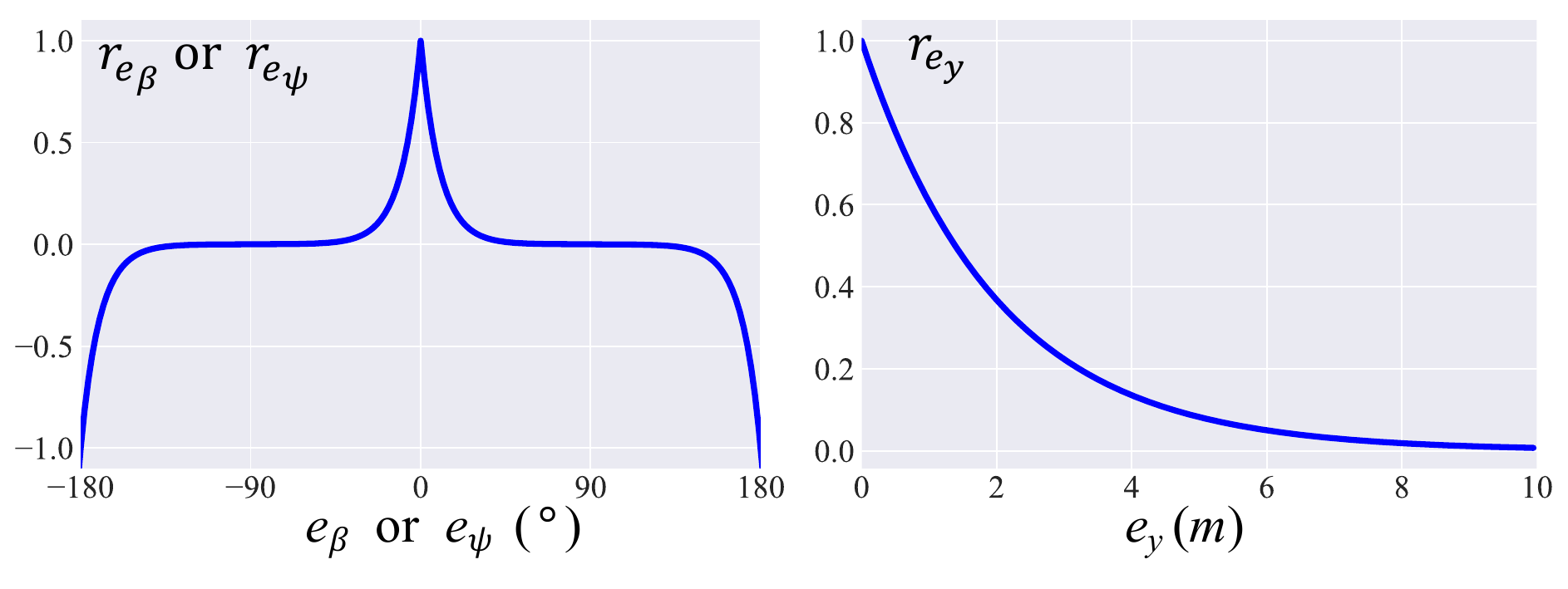}
    \caption{The partial rewards designed for vehicle drift control. The rewards reach a maximum value when the corresponding error is equal to 0, and decrease as the error increases. When the course angle error $e_{\psi}$ is larger than 90$^{\circ}$, $r_{e_{\psi}}$ become negative to further indicate a bad control command and prevent the vehicle from driving in the opposite direction.}
    \label{plot_reward}
    \vspace{-0.4cm}
\end{figure}

Perot et al. \cite{perot2017end} successfully used RL to control a simulated racing car. However, we observe a shaky control output in their demonstrated video. To avoid this phenomenon, we impose continuity in the action ($\mathbf{a}_t$), by constraining the change of output with the deployed action in the previous step ($\mathbf{a}_{t-1}$). The action smoothing strategy is
\begin{equation}
\setlength{\abovedisplayskip}{3pt}
\setlength{\belowdisplayskip}{3pt}
    \mathbf{a}_t = K_1\mathbf{a}_t^{net} + K_2\mathbf{a}_{t-1},
    \label{smoothing_equation}
\end{equation}
where $\mathbf{a}_t^{net}$ is the action predicted by the network with state $\textbf{s}_t$. $K_1$ and $K_2$ are the tuning diagonal matrices to adjust the smoothing effect. The larger the value of $K_2$, the more similar $\mathbf{a}_t$ and $\mathbf{a}_{t-1}$, and the smoother the corresponding control effect. Note that $K_{i(11)}$ influences the steering angle and $K_{i(22)}$ influences the throttle. We empirically select the value of $[K_{i(11)},K_{i(22)}]$ from a range of $\left\{[0.1,0.9], [0.3,0.7], [0.5,0.5], [0.7,0.3], [0.9,0.1]\right\}$, and finally set $K_1,\ K_2$ as follows.
\begin{equation}
\setlength{\abovedisplayskip}{5pt}
\setlength{\belowdisplayskip}{5pt}
K_1 =
\begin{bmatrix}
0.1 & 0 \\
0 & 0.3 
\end{bmatrix}, \ \  K_2 = 
\begin{bmatrix}
0.9 & 0 \\
0 & 0.7 
\end{bmatrix}.
\label{diag_k}
\end{equation}

\subsubsection{Reward shaping}
A reward function should be well defined to evaluate the controller performance, based on the goal of high-speed drifting through corners with low related errors ($e_y, e_{\psi}, e_{\beta}$). Therefore, we first design some partial rewards $r_{e_y},r_{e_{\psi}},r_{e_{\beta}}$ as (\ref{partial_rewards}), and illustrate them in Fig. \ref{plot_reward}. 
\begin{equation}
\setlength{\abovedisplayskip}{3pt}
\setlength{\belowdisplayskip}{3pt}
    \begin{aligned}
        &r_{e_y} = e^{-k_1 e_y}  \\
        &r_{e_{\psi}},\ r_{e_{\beta}} = f(x) = \left\{
                        \begin{aligned}
                        &e^{-k_2 |x|} \     &|x| &< 90^\circ \\
                        &-e^{-k_2 (180^\circ - x)} \    &x  &\geq 90^\circ \\
                        &-e^{-k_2  (180^\circ + x)} \    &x  &\leq -90^\circ
                        \end{aligned}
                        \right.
    \end{aligned}
    \label{partial_rewards}
\end{equation}
Note that $r_{e_{\psi}}$ and $r_{e_{\beta}}$ have the same computational formulae, which is denoted as $f(x)$, with $x$ representing $e_{\psi}$ or $e_{\beta}$. $k_1$ and $k_2$ are selected as 0.5 and 0.1. The total reward is defined as (\ref{total_reward}), which is the product of the vehicle speed and the weighted sum of partial rewards:
\begin{equation}
\setlength{\abovedisplayskip}{4pt}
\setlength{\belowdisplayskip}{4pt}
    r = v(k_{e_y}r_{e_y} + k_{e_{\psi}}r_{e_\psi} + k_{e_{\beta}}r_{e_{\beta}}).
    \label{total_reward}
\end{equation}
Speed factor $v$ is used to stimulate the vehicle to drive fast. If $v$ is smaller than 6 $m/s$, the total reward is decreased by half as a punishment; otherwise, the reward is the original product. The weight variables $[k_{e_y}, k_{e_{\psi}},k_{e_{\beta}}] $ are set to [40,40,20]. We empirically select these values from a range of $\left\{ [4, 4, 2], [20,20,20], [40,40,20], [400,400,200] \right\}$.

\subsubsection{Soft actor-critic}
\begin{figure}[t]
    \centering
    \includegraphics[width = \columnwidth]{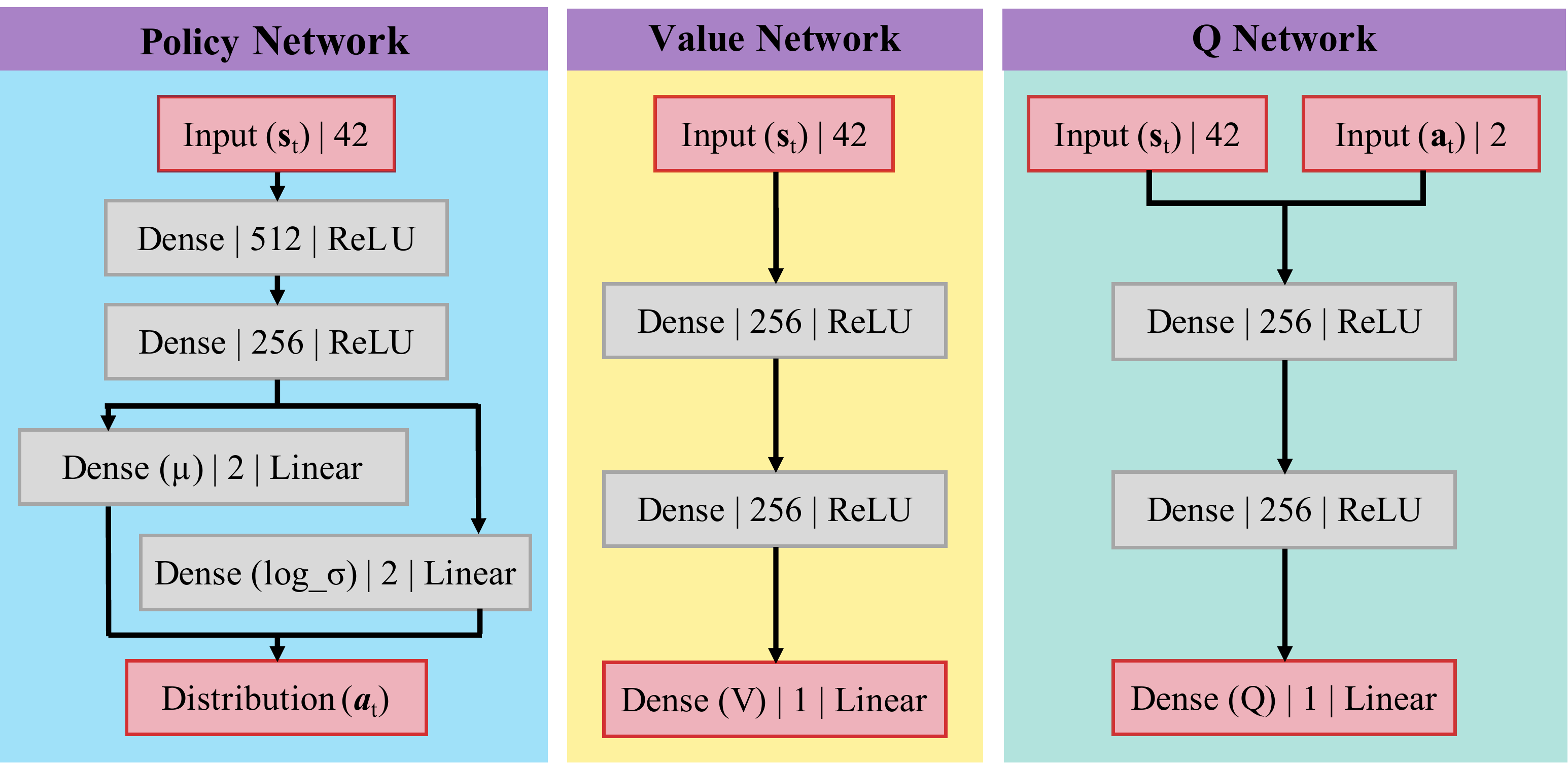}
    \caption{SAC network structures. The instructions in every layer indicate the network layer type, output channel dimension and activation function. \textit{Linear} here means no activation functions are used and \textit{Dense} means a fully connected neural network.}
    \label{networks}
    \vspace{-0.3cm}
\end{figure}

We choose SAC as our training algorithm, which optimizes a stochastic policy by maximizing the trade-off between the expected return and entropy with the off-policy learning method. It is based on the actor-critic framework, where the policy network is the actor, and the Q-network together with the value network is the critic. The critic can suggest a convergence direction for the actor to learn the optimal policy. In our experiments, three kinds of networks, including the policy network ($\pi_\upphi$), value network ($V_\uppsi$) and Q-networks ($Q_{\uptheta_1}$, $Q_{\uptheta_2}$) are learned. The different network structures are presented in Fig. \ref{networks}. In particular, two Q-networks with the same architecture are trained independently as the clipped double-Q trick, which can speed up training in this hard task, and the value network is used to stabilize the training. For more detailed information about the algorithm, we refer the reader to \cite{haarnoja2018soft}.

The complete training algorithm is shown in Algorithm \ref{alg:sac}. Firstly, the agent observes the current 42-dimensional state $\textbf{s}_t$, which is then transferred to a 2-dimensional action $\textbf{a}_t$ with fully-connected layers by the policy network. The action is sampled from the output distribution and normalized to $[-1,1]$ with the \textit{tanh} activation function. The sampled action is further mapped and smoothed to interact with the environment. When the agent obtains the next state $\textbf{s}_{t+1}$ and reward $r(\textbf{s}_t,\textbf{a}_t)$, the transition $\left( \textbf{s}_t,\textbf{a}_t,r(\textbf{s}_t,\textbf{a}_t),\textbf{s}_{t+1}\right)$ is stored into the replay buffer. Such interaction and stored procedures are repeated during training. At the end of the episodes, when the number of transitions is larger than the setting threshold, networks are updated respectively with the functions $J_{V}(\uppsi)$, $J_{Q}(\uptheta_1)$, $J_{Q}(\uptheta_2)$ and $J_{\pi}(\upphi)$, which are the same as those defined in \cite{haarnoja2018soft}. The whole procedure is repeated until the optimal policy is learned.

\section{Experiments and Discussion}
\subsection{Training setup}

\subsubsection{Implementation}

We train our \texttt{SAC} controller on six maps (Fig. \ref{maps} (a-f)). Map (a) is relatively simple and is used for the first-stage training, in which the vehicle learns some basic driving skills such as speeding up by applying large values of throttle and drifting through some simple corners. Maps (b-f) have different levels of difficulty with diverse corner shapes, which are used for further training with the pre-trained weights from map (a). The vehicle can use the knowledge learned from map (a) and quickly adapt to these tougher maps, to learn a more advanced drift technique. In this stage, maps (b-f) are randomly chosen for each training episode. In addition to the various road structures, we also hope the controller can handle other changing conditions. To this end, at the start of each episode, the tire friction and vehicle mass is sampled from the range of $[3.0,4.0]$ and $[1.7t,1.9t]$ respectively. Lower values make the vehicle more prone to slip, thus leading to a harder control experience. We use the Adam optimizer for training with a learning rate of 0.0003 and batch size of 512.

\begin{figure}[tpb]
    \centering
    \setlength{\abovecaptionskip}{-1pt}
    \includegraphics[width = 0.9\columnwidth]{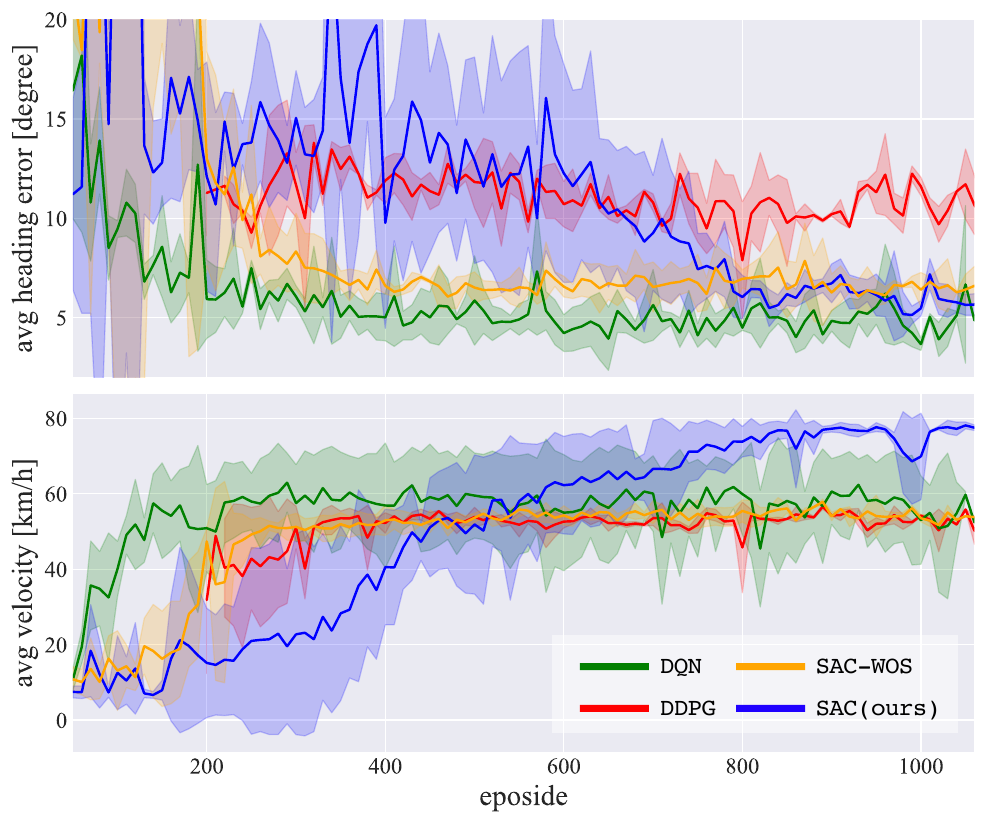}
    \caption{Performance curves of different algorithms during training on map (a). The plots are averaged over 3 repeated experiments. The solid curve corresponds to the mean, and the shaded region to the standard deviation. Note that the \texttt{DDPG} controller starts to be evaluated from the 200th episode, because the vehicle often gets stuck in circling around the start location in the early phase.}
    \label{simple_results}
    \vspace{-0.3cm}
\end{figure}

\subsubsection{Baselines}

For comparison, we train the controller with three other methods: 
\begin{itemize}
    \item \texttt{DQN}. Since it can only handle the discrete action space, we divide the range of the steering angle evenly for 10 values and throttle for 5 values. Thus, the number of candidate actions is 50 without the action smoothing strategy.
    \item \texttt{DDPG}. For better performance of this method, we set $K_{1(11)} = 0.6$ and $K_{2(11)} = 0.4$ in (\ref{diag_k}).
    \item \texttt{SAC-WOS}. We use SAC to train the controller but without the action smoothing strategy.
\end{itemize}

\subsubsection{Performance during training}



Fig. \ref{simple_results} shows the average heading angle error and the average speed of evaluation rollouts during training for \texttt{DQN}, \texttt{DDPG}, \texttt{SAC-WOS} and \texttt{SAC}. The results show that all methods can learn to speed up and reduce the error during training, and finally converge to their optimal values. In the end, they have approximately the same heading angle error, except for \texttt{DDPG}. However, \texttt{SAC} achieves a much higher average velocity (80 $km/h$) than the baselines. This illustrates that the \texttt{SAC} controller is capable of making the vehicle follow the reference trajectory accurately as well as maintain a high speed. In addition, it is shown that the action smoothing strategy can improve the final performance by comparing \texttt{SAC-WOS} and \texttt{SAC}.

\begin{table*}[t]
\newcommand{\tabincell}[2]{\begin{tabular}{@{}#1@{}}#2\end{tabular}}
\newcommand{\NA}{---}
        \renewcommand{\arraystretch}{1.3}
        \caption{Quantitative Evaluation and Generalization for Different Methods under Varied Environment Setups. $\uparrow$ Means Larger Numbers Are Better, $\downarrow$ Means Smaller Numbers Are Better. The Bold Font Highlights the Best Results in Each Column.}
        \label{evaluation_table}
        \centering
        \begin{tabular}{ c  c  c c c c c c  c c c c }
        \toprule
        {}&
        \multirow{3}{*}{Setup}&
        \multirow{3}{*}{Methods}& 
        \multicolumn{5}{c}{{Performance over the whole track}}&
        \multicolumn{4}{c}{{Performance through corners}}\\
        \cmidrule(lr){4-8} \cmidrule(lr){9-12}
        {}&
        \multirow{3}{*}{}&
        \multirow{3}{*}{} 
        &  {C.T.E.}$\downarrow$   & {H.A.E.}$\downarrow$ & {MAX-VEL}$\uparrow$ & {L.T.}$\downarrow$  & {SMOS}$\downarrow$ &  {C.T.E.}$\downarrow$   & {H.A.E.}$\downarrow$ & {AVG-VEL}$\uparrow$ & {SLIP}$\uparrow$ \\
        {}&
        \multirow{3}{*}{}&
        \multirow{3}{*}{}
        &  $(m)$  & $(^\circ)$ & $(km/h)$ & $(s)$ &  & $(m)$& $(^\circ)$ & $(km/h)$ & $(^\circ)$\\ 
        \midrule
        \multicolumn{1}{c|}{
        \multirow{12}{*}{\rotatebox{90}{\textbf{\textit{Evaluation}}}}}&
        \multirow{4}{*}{\tabincell{c}{F3.0M1.7}}&
        \texttt{DQN} &1.286 & 7.971 & 98.57 & 150.89 & 0.132 & 1.413 & 3.127 & 77.49 & 19.48\\
        
        \multicolumn{1}{c|}{
        \multirow{15}{*}{}}&
        \multirow{4}{*}{}&  
        \texttt{DDPG} &2.036 & 16.664 & 90.59 & 215.45 & 0.401 & 2.674 & 4.106 & 52.52 & 29.01\\
        \multicolumn{1}{c|}{
        \multirow{15}{*}{}}&
        \multirow{4}{*}{}&
        \texttt{SAC-WOS} &\textbf{0.811} & 6.703 & 84.02 & 186.12 & 0.632 & \textbf{1.030}& 4.559 & 60.82 & 23.84\\
        \multicolumn{1}{c|}{
        \multirow{15}{*}{}}&
        \multirow{4}{*}{}&
        \texttt{SAC} &0.900 & \textbf{5.707} & \textbf{103.71} & \textbf{145.90} & \textbf{0.130} & 1.174 & \textbf{3.009} & \textbf{78.85} & \textbf{29.23} \\
        \cline{3-12}
        
        \multicolumn{1}{c|}{
        \multirow{15}{*}{}}&
        \multirow{4}{*}{\tabincell{c}{F3.5M1.8}}&
        \texttt{DQN} &1.125 & 7.118 & 96.45 & 149.34 & 0.131 & 1.277 & 3.957 & 76.85 & 24.12\\
        
        \multicolumn{1}{c|}{
        \multirow{15}{*}{}}&
        \multirow{4}{*}{}&  
        \texttt{DDPG} &1.805 & 16.921 & 84.27 & 216.25 & 0.402 & 2.178 & 4.556 & 52.21 & 25.28\\
        
        \multicolumn{1}{c|}{
        \multirow{15}{*}{}}&
        \multirow{5}{*}{}&
        \texttt{SAC-WOS} &\textbf{0.734} & 6.434 & 84.04 & 187.57  & 0.642 & \textbf{0.970} & 4.342 & 60.49 & 24.76\\
        \multicolumn{1}{c|}{
        \multirow{17}{*}{}}&
        \multirow{5}{*}{}&
        \texttt{SAC} &0.907 & \textbf{5.776} & \textbf{103.45} & \textbf{143.42} & \textbf{0.125} & 1.361 & \textbf{2.129} & \textbf{79.07} & \textbf{26.17} \\
        
        \cline{3-12}
        
        \multicolumn{1}{c|}{
        \multirow{17}{*}{}}&
        \multirow{4}{*}{\tabincell{c}{F4.0M1.9}}&
        \texttt{DQN} &1.114 & 6.943 & 96.97 & 149.08 & 0.131 & 1.330 & 3.918 & 76.74 & 20.76\\
        \multicolumn{1}{c|}{
        \multirow{17}{*}{}}&
        \multirow{5}{*}{}&  
        \texttt{DDPG} &1.629 & 15.899 & 82.59 & 212.94 & 0.402 & 1.897 & 4.005 & 52.94 & 21.62\\
        \multicolumn{1}{c|}{
        \multirow{17}{*}{}}&
        \multirow{5}{*}{}&
        \texttt{SAC-WOS} &\textbf{0.736} & 6.170 & 81.12 & 191.25 & 0.655 & \textbf{1.006} & 4.064 & 59.07 & 23.85\\
        \multicolumn{1}{c|}{
        \multirow{17}{*}{}}&
        \multirow{5}{*}{}&
        \texttt{SAC} &0.920 & \textbf{5.850} & \textbf{102.78} & \textbf{142.44}& \textbf{0.123} & 1.526 & \textbf{1.691} & \textbf{78.93} & \textbf{24.28}\\

        \hline
        \multicolumn{1}{c|}{
        \multirow{6}{*}{\rotatebox{90}{\textbf{\textit{Generalization}}}}}&
        F2.6M1.6&
        \multirow{6}{*}{\texttt{SAC}}&
        1.219 & 6.757 & 105.69 & 148.87 & 0.132 & 1.346 & 3.443 & 79.46 & 43.27\\
        \multicolumn{1}{c|}{
        \multirow{6}{*}{}}&
        F4.4M2.0&
        \multirow{6}{*}{}&
        1.001 & 6.351 & 100.83 & 143.21 & 0.124 & 1.701 & 1.477 & 78.20 & 23.08\\
        \multicolumn{1}{c|}{
        \multirow{6}{*}{}}&
        DF-M1.8&
        \multirow{6}{*}{}&
        1.021 & 6.718 & 102.46 & 144.70 & 0.129 & 1.186 & 2.748 & 80.13 & 12.54\\
        
        \multicolumn{1}{c|}{
        \multirow{6}{*}{}}&
        Vehicle-2&
        \multirow{6}{*}{}&
        1.000 & 6.252 & 128.07 & 126.28 & 0.140 & 1.345 & 1.382 & 92.77 & 45.03\\
        \multicolumn{1}{c|}{
        \multirow{6}{*}{}}&
        Vehicle-3&
        \multirow{6}{*}{}&
        0.918 & 6.368 & 123.75 & 126.55 & 0.138 & 1.322 & 1.687 & 91.23 & 39.38\\
        
        \multicolumn{1}{c|}{
        \multirow{6}{*}{}}&
        Vehicle-4&
        \multirow{6}{*}{}&
        0.450 & 3.486 & 67.46 & 187.94 & 0.103 & 0.659 & 1.514 & 59.46 & 9.74\\
        
        \hline
        \multicolumn{1}{c|}{
        \multirow{3}{*}{\rotatebox{90}{\textbf{\textit{AppTest}}}}}&
        \multirow{3}{*}{F3.5M1.8}&
        \texttt{HUMAN-DFT}&
        \NA & \NA & {112.86} & 141.69 & 0.055 & \NA & \NA & 79.38 & 28.59\\
        \multicolumn{1}{c|}{
        \multirow{3}{*}{}}&
        \multirow{3}{*}{}&
        \texttt{HUMAN-NORM}&
        \NA & \NA & 108.58 & 160.53 & {0.011} & \NA & \NA & 66.57 & 8.28\\
        \multicolumn{1}{c|}{
        \multirow{3}{*}{}}&
        \multirow{3}{*}{}&
        \texttt{SAC-APP}&
        \NA & \NA & 105.59 & 143.59 & 0.106 & \NA & \NA & 80.30 & 26.87 \\ 
        
        \bottomrule
        \end{tabular}
        \vspace{-0.2cm}
\end{table*}


\subsection{Evaluation}
To evaluate the controller performance, we select three combinations of tire friction (F) and vehicle mass (M) as F3.0M1.7, F3.5M1.8 and F4.0M1.9. The test environment is an unseen tough map (g) with various corners of angles ranging from 40$^{\circ}$ to 180$^{\circ}$. 


\subsubsection{Performance metrics}
We adopt seven metrics to measure the performance of different methods.
\begin{itemize}
    \item \textbf{C.T.E.} and \textbf{H.A.E.} is the cross track error and heading angle error, respectively.
    \item \textbf{MAX-VEL} and \textbf{AVG-VEL} is the maximum and average velocity of a driving test, respectively.
    \item \textbf{L.T.} is the time to reach the destinations (the lap time).
    \item \textbf{SMOS} measures the smoothness of driving, calculated by the rolling standard deviation of steering angles during a driving test.
    \item \textbf{SLIP} is the maximum slip angle during a driving test. Since larger slip angles mean larger usable state spaces beyond the handling limits, it can indicate a more powerful drift controller.
\end{itemize}

\subsubsection{Quantitative results}

\begin{figure*}[tpb]
    \centering
    \setlength{\abovecaptionskip}{-0.5pt}
    \includegraphics[width = 2.05\columnwidth]{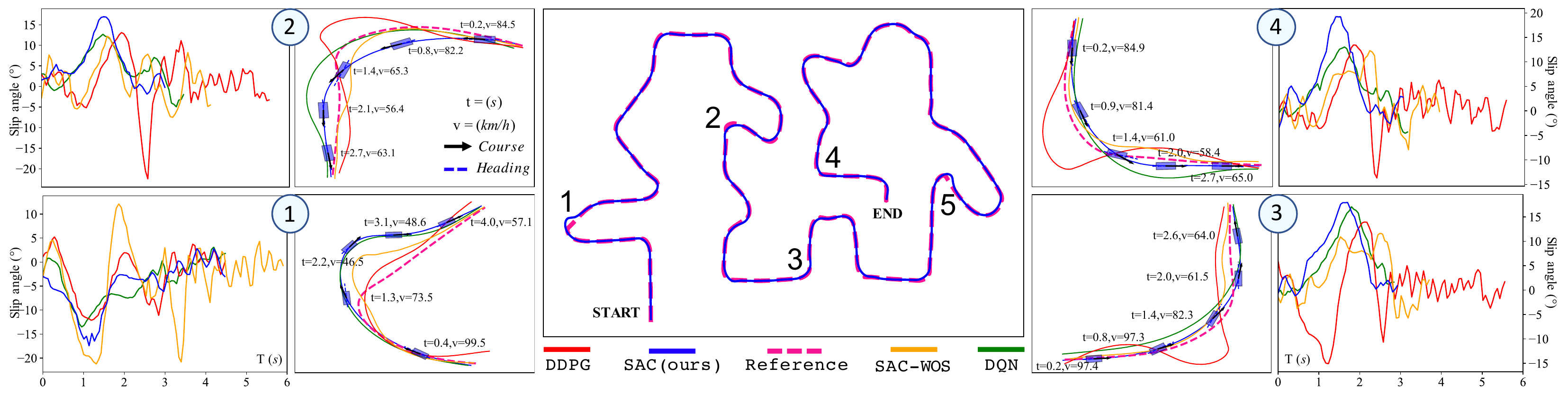}
    \caption{Qualitative trajectory results on map (g) based on the setup F3.5M1.8. The picture in the middle represents the overall trajectory of the human driver (i.e., \texttt{Reference}) and our \texttt{SAC} controller. The pictures on either side depict some drift-cornering trajectories and corresponding slip angle curves from different controllers. For further analysis of the \texttt{SAC} controller, we label some state information over time along these trajectories, which are velocity, moving direction (course) and heading direction. Note that the difference between the course and the heading is the slip angle.}
    \label{trajectory}
    \vspace{-0.3cm}
\end{figure*}


All controllers are tested four times on map (g) and the average evaluation results are presented in Table \ref{evaluation_table}. Apart from the overall performance through the whole track, the results for driving through corners are also listed, to give a separate analysis on drift ability. Additionally, two reference results on F3.5M1.8 from the human driver are presented for comparison, in which \texttt{HUMAN-DFT} drifts through sharp corners and \texttt{HUMAN-NORM} slows down and drives cautiously through corners.



\textbf{Time cost and velocity}. Our \texttt{SAC} controller achieves the shortest lap time in all setups, with the maximum velocity among the four methods. In particular, the speed reaches up to 103.71 $km/h$ in setup F3.0M1.7, which is much higher than \texttt{DDPG} (90.59 $km/h$) and \texttt{SAC-WOS} (84.02 $km/h$). Compared with \texttt{HUMAN-NORM}, the \texttt{SAC} controller adopts the drifting strategy, which achieves a much shorter lap time.


\textbf{Error analysis}. C.T.E. and H.A.E. indicate whether the vehicle can follow the reference trajectory accurately. The \texttt{SAC-WOS} controller achieves the smallest C.T.E., but \texttt{SAC} is the best for H.A.E., especially through corners. The possible reason is \texttt{SAC} controls the car to drift through corners with similar slip angles to the reference behaviors, while other methods tend to mainly track the positions on the trajectory.

\begin{figure}[tpb]
    \centering
    \setlength{\abovecaptionskip}{-0.5pt}
    \includegraphics[width = 0.9\columnwidth]{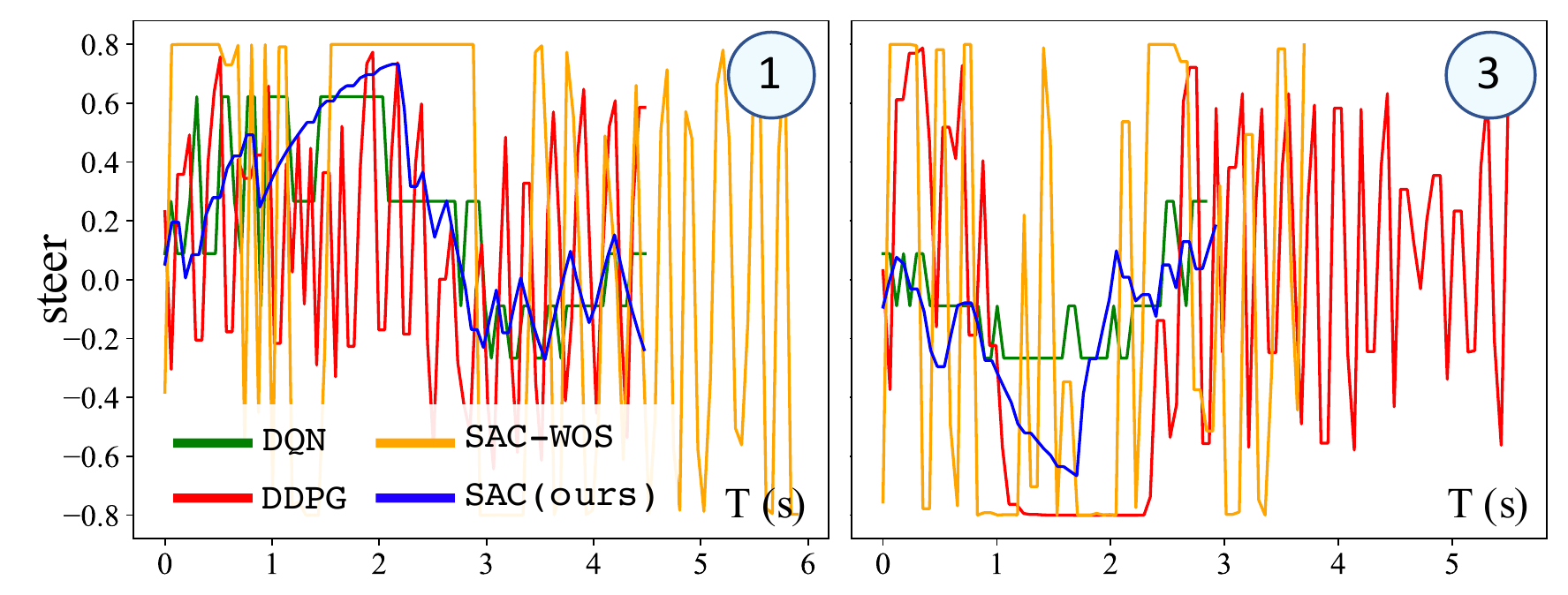}
    \caption{The curves of steering control command through corner-1 and corner-3 from different controllers.}
    \label{cor_steer}
    \vspace{-0.4cm}
\end{figure}

\textbf{Drifting velocity and slip angle}. We calculate the average velocity and the largest slip angle while drifting. In all setups, the \texttt{SAC} controller achieves the highest speed and largest slip angles. In setup F3.5M1.8, the AVG-VEL reaches up to 79.07 $km/h$, which is very similar to \texttt{HUMAN-DFT} (79.38 $km/h$). In setup F3.0M1.7, the SLIP of the \texttt{SAC} controller reaches up to 29.23$^{\circ}$, which is much higher than \texttt{DQN} and \texttt{SAC-WOS}. On the other hand, although the \texttt{DDPG} and \texttt{SAC-WOS} controller can generate large slip angles, their control outputs are rather shaky, leading to velocities even lower than \texttt{HUMAN-NORM}.


\textbf{Driving smoothness}. SMOS reflects how steady the vehicle is while driving. Although all controllers generate larger values of SMOS than the human driver, \texttt{SAC} achieves the smallest among them.




\begin{center}
    \begin{table}[t]
        \renewcommand{\arraystretch}{1.3}
        \caption{Vehicles Used for Training and Testing Our Model. The Vehicle Used for Training is Boldfaced. MOI is the Moment of Inertia of the Engine around the Axis of Rotation.}
        \label{vehicle_table}
        \centering
        \begin{tabular}{c  c  c c c}
        \toprule
        \multirow{2}{*}{Physics}&
        \textbf{Vehicle-1} & Vehicle-2 & Vehicle-3 & Vehicle-4 \\
        \multirow{2}{*}{}&
        \textbf{\textit{(Audi A2)}} & \textit{(Audi TT)} &\textit{(Citroen C3)}  & \textit{(ColaTruck)}\\
        \midrule
        Mass $(t)$ & 1.7-1.9  & 1.2 & 1.0 &5.5\\
        Tire friction & 3-4  & 3.5 & 3.5 &3.5\\
        MOI $(kgm^2)$ & 0.1  & 0.3 & 1.0 &1.0 \\
        Length $(m)$ & 3.6  & 4.2& 3.8 &5.6\\
        Width $(m)$ & 1.6  & 1.8 & 1.7 &2.5\\
        Drive type & FWD &FWD&FWD &AWD\\
        \bottomrule
        \end{tabular}
    \end{table}
    \vspace{-0.3cm}
\end{center}

\subsubsection{Qualitative results}

Fig. \ref{trajectory} shows the qualitative trajectory results on the test map (g). The \texttt{SAC} controller is shown to have excellent performance in tracking the trajectory on linear paths and most of the corners. Some mismatches may occur if the corner angle is too small (e.g., <50$^\circ$), such as corner-1 and corner-5. In corner-3 and corner-4 with angles of about 90$^\circ$, the drift trajectory of our \texttt{SAC} controller is very similar to that of the reference, even though the speed at the entry of corner-3 is near 100 $km/h$. During the drifting process of \texttt{SAC}, the speed is reduced when the vehicle drives close to the corner vertex, resulting from the large slip angle. However, it still maintains a high speed and accelerates quickly when leaving the corner. 

Since all controllers output the largest value of throttle to maintain a high speed, except DQN (which jumps between 0.9 and 1.0), we plot their steering curves for corner-1 and corner-3 for further comparison. These are presented in Fig. \ref{cor_steer}. It is shown that the steering angles of the other controllers are tremendously shaky, especially for \texttt{DDPG} and \texttt{SAC-WOS}. In contrast, the steering angle of \texttt{SAC} controller concentrates in a smaller range and is much smoother.

\begin{figure}[tpb]
    \centering
    \includegraphics[width = 0.8\columnwidth]{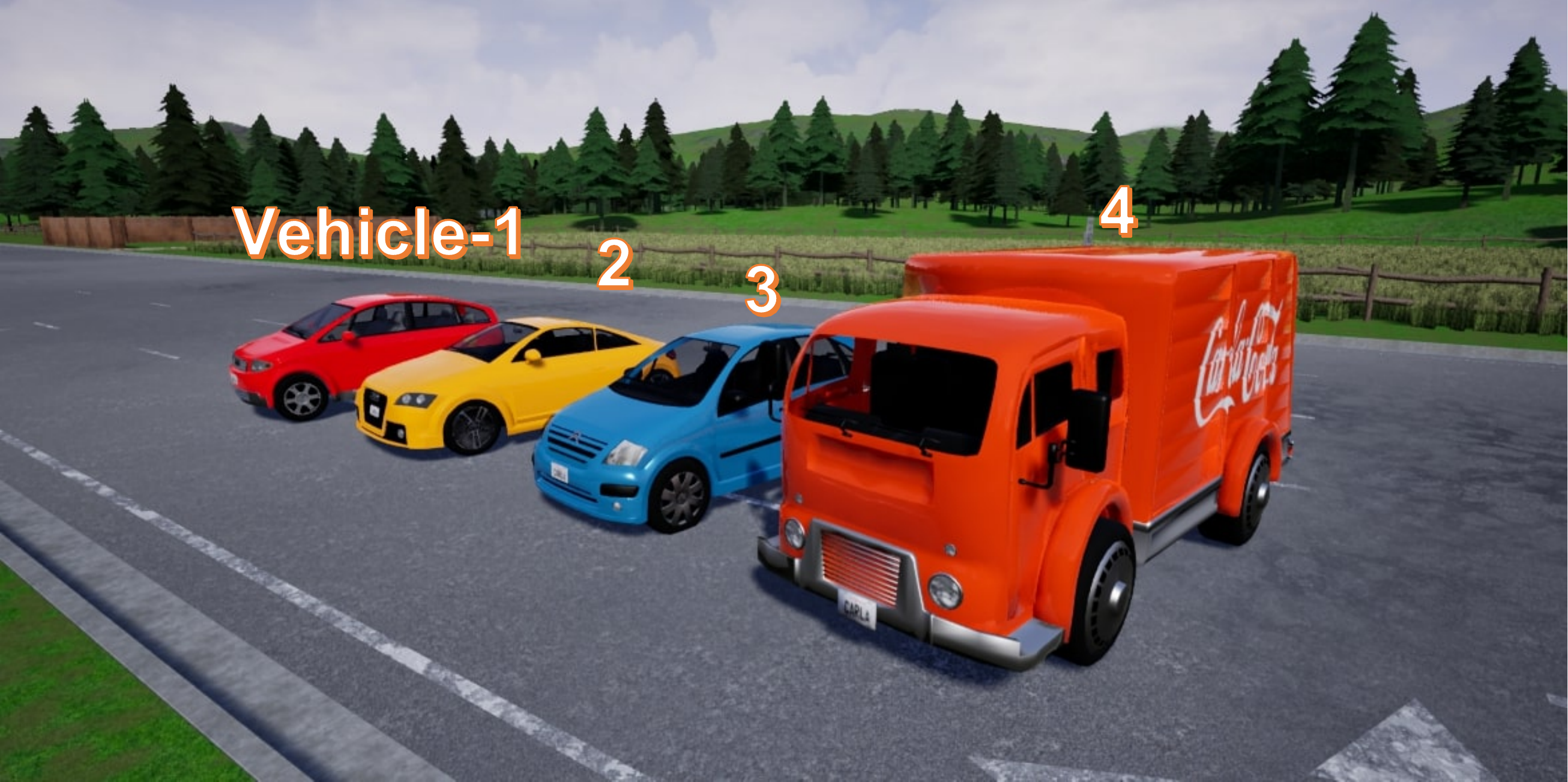}
    \caption{Vehicles used for training and testing our model.}
    \label{vehicles_photo}
    \vspace{-0.3cm}
\end{figure}

\subsection{Generalization}
To test the generalization ability of the proposed \texttt{SAC} controller, we evaluate it with varied tire friction, vehicle mass and vehicle types on map (g). Different vehicles and their physics are shown in Fig. \ref{vehicles_photo} and Table \ref{vehicle_table}. The performance results are presented in Table \ref{evaluation_table}-Generalization.
\subsubsection{Unseen mass and friction}
We set two combinations of unseen friction and mass on vehicle-1 as F2.6M1.6 and F4.4M2.0. Our \texttt{SAC} controller can handle them without any fine-tuning and the highest speed is more than 100 $km/h$. Drifting is completed successfully and the maximum slip angle is up to 43.27$^\circ$ for F2.6M1.6. Additionally, we test the proposed controller using vehicle-1 with different tire friction in the front wheels (2.8) and rear wheels (4.2). This is called DF-M1.8, since sometimes wear conditions vary on different tires for a vehicle. In this tough setup, our \texttt{SAC} controller can make the vehicle drive through the whole map fast and smoothly. However, the drift control performance does not meet with expectations, with the maximum slip angle smaller than 20$^\circ$. This is caused by the large rear tire friction, which makes it difficult for the car to slip.

\subsubsection{Unseen vehicle types}
The \texttt{SAC} controller is further tested by three other types of vehicles. Vehicle-2 is similar to Vehicle-1 but is about 0.5$t$ lighter, Vehicle-3 has a much larger MOI and bigger mass, and Vehicle-4 is an all-wheel drive heavy truck with distinct physical parameters. The results show that our \texttt{SAC} method achieves a notable generalization performance on these unseen vehicles. For Vehicle-2, the highest speed is up to 128.07 $km/h$, and the average drift speed is 92.77 $km/h$, both of which are even better than Vehicle-1 with the benefit of a smaller mass and a more powerful engine. The same is true of Vehicle-3. For Vehicle-4, the \texttt{SAC} controller is sufficiently capable of controlling it to follow a reference trajectory precisely, but the drift performance is not satisfactory with small slip angles and cornering speeds, due to its heavy weight and large size. Note that for each kind of vehicle, the referenced drift trajectories are different in order to meet the respective physical dynamics.

\subsubsection{Application test without expert reference}
To evaluate whether the proposed \texttt{SAC} model can be deployed in scenarios where expert driving trajectories are not available, we further test it by providing less information. In CARLA, the $(x,y)$ waypoints in the center of the road can easily be obtained, so they are used to form a rough reference trajectory. The directions of this trajectory $\psi^{ref}$ are derived based on its tangents for calculating the heading angle errors $e_{\psi}$, and the reference slip angles are set to zero. Accordingly, $e_{\beta}, \dot e_{\beta}, e_{vy}$ and $\dot e_{vy}$ in the state space are also set to zero. The reference forward velocities are set to 110 $km/h$ constant along the whole track. Based on this setup, Vehicle-1 of F3.5M1.8 is tested on map (g) and the results are shown in Table \ref{evaluation_table}-AppTest as \texttt{SAC-APP}. It is very interesting that although a rough trajectory is used, the final performance is still comparable with the \texttt{SAC} controller provided with accurate expert trajectories. 

\begin{table*}[t]
\newcommand{\tabincell}[2]{\begin{tabular}{@{}#1@{}}#2\end{tabular}}
\newcommand{\NA}{---}
        \setlength{\abovecaptionskip}{-0.5pt}
        \renewcommand{\arraystretch}{1.3}
        \caption{Quantitative Evaluation for Policies Trained with (\texttt{SAC-42}) or without (\texttt{SAC-30}) Slip Angle as Guidance. $\uparrow$ Means Larger Numbers Are Better, $\downarrow$ Means Smaller Numbers Are Better. The Bold Font Highlights the Best Results in Each Column.}
        \vspace{0.1cm}
        \label{tab:slip-angle}
        \centering
        \begin{tabular}{c  c  c c c c c c  c c c c}
        \toprule
        \multirow{3}{*}{Setup}&
        \multirow{3}{*}{Methods}& Training &
        \multicolumn{5}{c}{{Performance over the whole track}}&
        \multicolumn{4}{c}{{Performance through corners}} \\
        \cmidrule(lr){3-3} \cmidrule(lr){4-8} \cmidrule(lr){9-12}
        
        \multirow{3}{*}{}&
        \multirow{3}{*}{} & Time $\downarrow$ &  {C.T.E.}$\downarrow$   & {H.A.E.}$\downarrow$ & {MAX-VEL}$\uparrow$ & {L.T.}$\downarrow$  & {SMOS}$\downarrow$ & C.T.E. $\downarrow$& H.A.E. $\uparrow$ & {AVG-VEL}$\uparrow$ & {SLIP}$\uparrow$ \\
        
        \multirow{3}{*}{} & \multirow{3}{*}{}
        & $(hours)$ & $(m)$  & $(^\circ)$ & $(km/h)$ & $(s)$ & &$(m)$ &$(^\circ)$ & $(km/h)$ & $(^\circ)$ \\ 
        \midrule
        \multirow{2}{*}{F3.5M1.8}&
        \texttt{SAC-42} &\textbf{11.03} & \textbf{0.896} & \textbf{5.390} & \textbf{122.44} & \textbf{48.01} & \textbf{0.094}  & \textbf{1.298} & \textbf{5.992} & 86.06 & \textbf{32.85} \\
        
        \multirow{2}{*}{}&
        \texttt{SAC-30} &23.47 & 1.390 & 7.908 & 114.97 & 51.60& 0.106 &2.671 &15.943 & \textbf{86.75} & 28.47\\
        \bottomrule
        \end{tabular}
        \vspace{-0.3cm}
\end{table*}

Since we mainly exclude the information of slip angle here, it can be inferred that they are dispensable for the policy execution in our task. This phenomenon is valuable, indicating that our drift controller could be applied to unseen tracks without generating an accurate reference trajectory in advance. This is critical for further real-world applications where a rough reference could be derived online from 2D or 3D maps, which are common in robot applications.

\subsection{Ablation study}
We have shown above that a rough trajectory is sufficient for the application. Therefore, can we also provide less information during the training and achieve no degradation in the final performance? To answer this question, a comparison experiment on map (a) is conducted by training an additional controller excluding variables related to the slip angle in the reward and state space ($e_{\beta},\dot e_{\beta}$ and 10 reference slip angles in $\mathcal{T}$). In this way the state space becomes 30-dimensional. Accordingly, the corresponding controller is named \texttt{SAC-30}, and \texttt{SAC-42} indicates the original one. These controllers are tested four times and the average evaluation results are presented in Table \ref{tab:slip-angle}. It shows that \texttt{SAC-42} costs much less training time but achieves better performance with a higher speed, shorter lap time and smaller error. It also drives more smoothly than \texttt{SAC-30}. Generally, accurate slip angles from expert drift trajectories are indeed necessary in the training stage, which can improve the final performance and the training efficiency.

\section{Conclusion}
In this paper, to realize high-speed drift control through manifold corners for autonomous vehicles, we propose a closed-loop controller based on the model-free deep RL algorithm soft actor-critic (\texttt{SAC}) to control the steering angle and throttle of simulated vehicles. The error-based state and reward are carefully designed and an action smoothing strategy is adopted for stable control outputs. Maps with different levels of driving difficulty are also designed to provide training and testing environments.

After the two-stage training on six different maps, our \texttt{SAC} controller is sufficiently robust against varied vehicle mass and tire friction to drift through complex curved tracks quickly and smoothly. In addition, its remarkable generalization performance has been demonstrated by testing different vehicles with diverse physical properties. Moreover, we have discussed the necessity of slip angle information during training, and the non-degraded performance with a rough and easy-to-access reference trajectory during testing, which is valuable for applications. To reduce the labor costs in generating accurate references for training, we will explore leaning-based methods for trajectory planning in drift scenarios, which is left to future work.

\bibliographystyle{IEEEtran}
\bibliography{main.bib}






\end{document}